\documentclass{article} 
\usepackage{iclr2024_conference, times}

\usepackage{amsmath,amsfonts,bm}

\def\eqref#1{equation~\ref{#1}}

\def\1{\bm{1}}

\DeclareMathAlphabet{\mathsfit}{\encodingdefault}{\sfdefault}{m}{sl}
\SetMathAlphabet{\mathsfit}{bold}{\encodingdefault}{\sfdefault}{bx}{n}

\usepackage{hyperref}
\usepackage{url}

\usepackage{graphicx}
\usepackage{multirow}
\usepackage{booktabs}
\usepackage{mathrsfs}
\usepackage{tcolorbox}
\usepackage{wrapfig}

\title{Evaluating Tool-Augmented Agents in Remote Sensing Platforms}

\author{Simranjit Singh, Michael Fore, Dimitrios Stamoulis \\
\textit{CoStrategist} R\&D Group, Microsoft Corporation, Redmond, WA, USA \\
\texttt{\{simsingh, mifore, stamoulis.dimitrios\}@microsoft.com}
}

\iclrfinalcopy 
\begin{document}

\maketitle

\begin{abstract}
Tool-augmented Large Language Models (LLMs) have shown impressive capabilities in remote sensing (RS) applications. However, existing benchmarks assume question-answering input templates over predefined image-text data pairs. These standalone instructions neglect the intricacies of realistic \textit{user-grounded} tasks. Consider a geospatial analyst: they zoom in a map area, they draw a region over which to collect satellite imagery, and they succinctly ask ``\textit{Detect all objects here}''. Where is \textit{here}, if it is not explicitly hardcoded in the image-text template, but instead is implied by the system state, \textit{e.g.}, the \textit{live} map positioning? To bridge this gap, we present \texttt{GeoLLM-QA}, a benchmark designed to capture long sequences of verbal, visual, and click-based actions on a \textit{real} UI platform. Through in-depth evaluation of state-of-the-art LLMs over a diverse set of 1,000 tasks, we offer insights towards stronger agents for RS applications.
\end{abstract}

\section{Introduction}
\begin{figure}[h!]
    \centering    
    \includegraphics[width=1.0\linewidth]{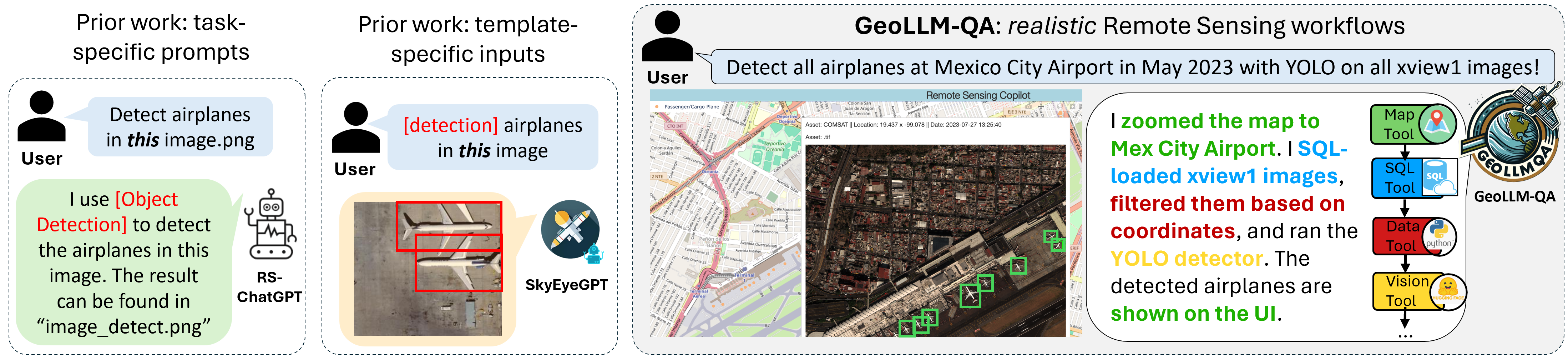}
    \vspace{-20pt}
    \caption{Unlike prior work that assumes task-specific templates (\textit{e.g.}, ``[detection]'' keyword), \texttt{GeoLLM-QA} requires the agent to follow nuanced instructions and perform multi-step reasoning to accomplish user-defined objectives.}
    \label{fig:geoflow}
\end{figure}

Large Language Models (LLMs) demonstrate impressive potential in complex geospatial scenarios, augmenting remote sensing (RS) platforms with agents capable of sophisticated planning, reasoning, and task execution. These developments have sparked interest to deploy multimodal models across various RS tasks, including image captioning and visual question answering (VQA)~\citep{Yuan_2022}. Notably, SkyEyeGPT~\citep{zhan2024skyeyegpt} finetunes state-of-the-art VQA agents~\citep{chen2023minigptv2} on RS imagery for unified multimodal responses, while Remote Sensing ChatGPT~\citep{guo2024remotesensinggpt} deploys computer-vision models (\textit{e.g.}, land use classification, object detection) via prompting. However, these approaches rely on chatbot-based templates with predefined text-image correlations over specific image files to assess LLM performance (Fig.~\ref{fig:geoflow} left), hence failing to capture the nuances of realistic \textit{user-grounded} RS tasks.

In this work, we aim to bridge this gap with the following contributions: \textit{\textbf{first}}, we introduce \texttt{GeoLLM-QA}, a novel benchmark of \textbf{1,000 diverse tasks}, designed to capture complex RS workflows where LLMs handle complex data structures, nuanced reasoning, and interactions with dynamic user interfaces (Fig.~\ref{fig:geoflow} right). To this end, we harness recent advancements in benchmarking work for tool-augmented LLMs~\citep{zhuang2023toolqa, maini2024tofu, koh2024visualwebarena}. \textit{\textbf{Second}}, we adopt a comprehensive evaluation scheme~\citep{maini2024tofu} beyond traditional text-based metrics that accurately assesses an agent's proficiency in utilizing external tools for effective problem-solving. \textit{\textbf{Third}}, we evaluate several state-of-the-art tool-augmentation and prompting methodologies on our benchmark. We highlight our key takeaways regarding the strengths, weaknesses, and potential of LLMs within geospatial platforms. We strive to motivate future work and help the RS community in unlocking further advancements in this domain.

\begin{figure}
    \centering    
    \includegraphics[width=1.0\linewidth]{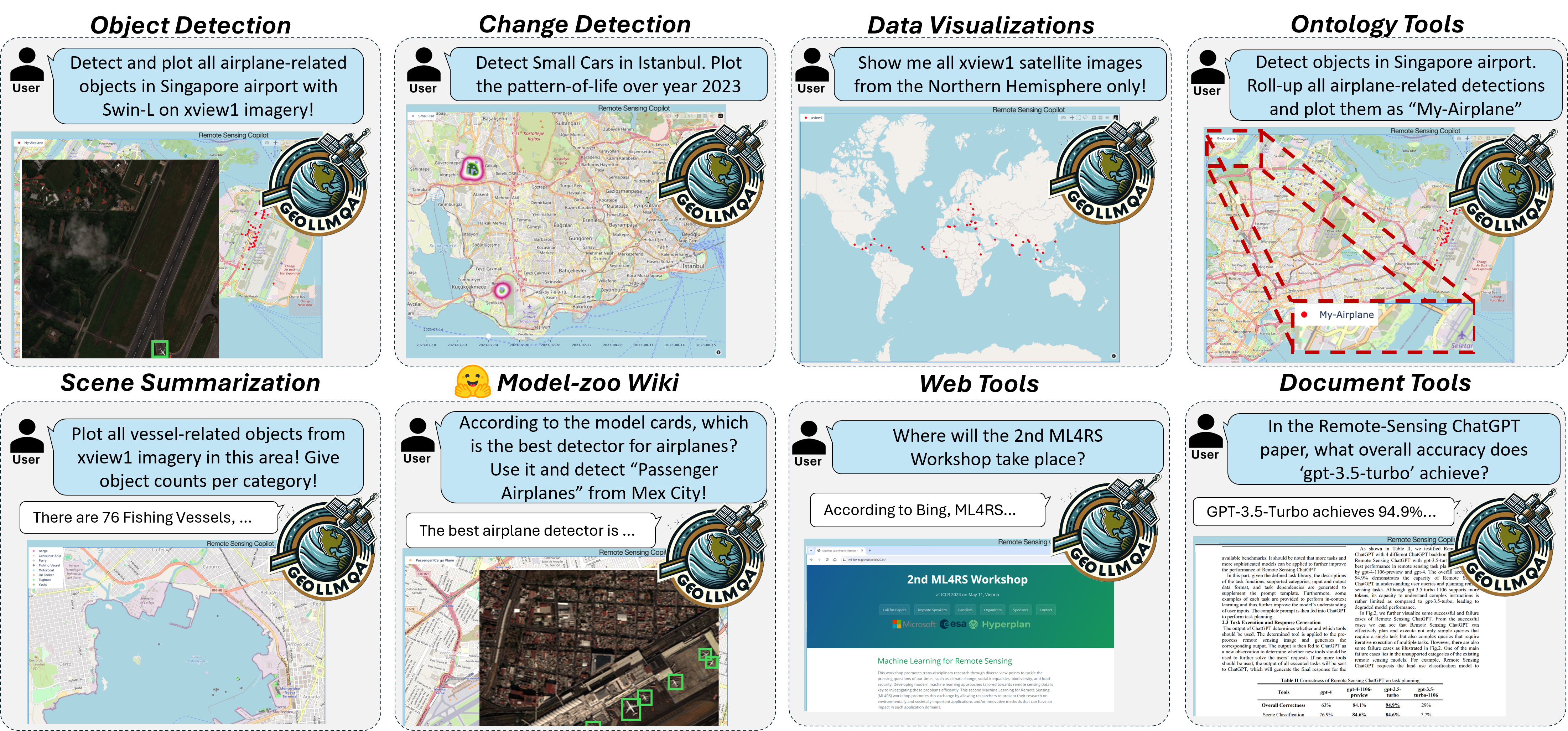}
    \caption{\texttt{GeoLLM-QA} challenges agents to solve complex RS tasks through multimodal reasoning and actions over long sequences of verbal, visual, and click-based actions on a \textit{real} UI platform.}
    \label{fig:scenarios}
\end{figure}

\section{The \texttt{GeoLLM-QA} Framework}

\textbf{Benchmarking Platform}: To assess geospatial reasoning in an agent-assisted \textit{platform} context, we draw inspiration from~\citep{zhou2023webarena} and we implement a benchmarking UI, as a realistic and reproducible standalone web-app that incorporates user-centered tasks with open-source tools and datasets. By leveraging open-source APIs, not only we address challenges of reproducibility and comparison across different systems, but also enable the examination of a wide range of RS use-cases through various input modalities including verbal, visual, and tactile interactions. The complete tool set consists of 117 tools, such as \texttt{plotly mapbox} APIs for the map functionality and \texttt{LangChain} routines for \texttt{FAISS} vectorstores~\citep{douze2024faiss}, to name a few. We intend to release our codebase and benchmark to stimulate future research on geospatial Copilots.

\textbf{Problem Formulation}: To denote RS tasks beyond simplistic VQA data-pairs, we model the problem after the realistic UI experience: intuitively, each interaction consists of the user question, the sequence of tool-calls by the agent, and the final (textual) response to user and platform state. We can therefore denote each task as $\{q, T, r, S\}$, where $q$ is the user prompt, $r$ is the textual response, while $T$ represents the set of tool-calling steps $T = \{t_1, t_2, \dots\}$. At each step $i$, the agent invokes tool $t_i= \{tool_i, args^{**}_i\} \in \mathcal{T}$ from the available tool space $\mathcal{T}$. Finally, $S$ defines the final system state: \textit{e.g.}, map positioning, loaded database, visible data holdings, \textit{etc}.

\textbf{Data Sources}: Our evaluation framework
includes three representative large-scale datasets: \texttt{xview1}~\citep{lam2018xview}, \texttt{xview3}~\citep{paolo2022xview3sar}, \texttt{DOTA-v2.0}~\citep{ding2021dota}. Encompassing both optical and synthetic aperture radar (SAR) imagery, these data holdings offer detailed object annotations across 80 categories from a total of 5,000 images. Notably, these datasets come with valuable metadata, such as dates and coordinates, which greatly enhances the complexity of temporal and spatial RS scenarios in our benchmark. The satellite imagery serves as \textit{task context} for LLM agents to execute function calls and is \textit{not} used for finetuning the LLM or other downstream tasks, enabling our research-purposes investigation. 

\textbf{``Golden'' Detector Models}: without loss of generality, we employ ``oracle detectors,'' a common practice in foundation-models literature~\citep{yang2023setofmark}, so that we can concentrate on evaluating the agent's proficiency in selecting and utilizing the appropriate tools without confounding the false positives/negatives of a non-optimal detector. By abstracting out detection errors, we can measure any degradation in performance metrics directly attributable to agents' failures.

For instance, consider a scenario where the LLM is instructed to ``\textit{detect all airplanes at the Mexico City airport using the YOLO detector}.'' We want to verify whether the agent can designate the right detector, filter through the correct imagery, and specify the right classes. Therefore, upon the LLM's selection of an image set, we assume an oracle detector that provides 100\% accurate detections, \textit{i.e.}, ``gold'' results directly from dataset ground truths. We then calculate the recall of these detector ``results'', attributing any discrepancies solely to the agent’s inability to accurately fulfill the task. 

\textbf{Benchmark Creation}: To create a representative set of RS tasks, \texttt{GeoLLM-QA} adopts the three-step benchmarking process presented in~\citep{zhuang2023toolqa}: \textit{1. Reference Template Collection}: we curate a set of 25 template questions that cover the wide range of RS tasks, such as object detection, change detection, \textit{etc}. Several key tasks are shown in Fig.~\ref{fig:scenarios}. To generate answers for these questions, we guide \texttt{GPT-4} to reach the answers via a simple human-in-the-loop mechanism via \textit{feedback} UI buttons~\citep{ouyang2022training}. By using previous (un)successful attempts as in-context examples, GPT can quickly help us create the Reference Templates.

\textit{2. LLM-guided Question Generation}: we generate permutations and perturbations of the Reference Templates. Note here that previous RS benchmarks assume that all LLM tasks are implicitly correct. However,  \citet{maini2024tofu} show that one of the most challenging aspects of agent performance is their ability to handle prompts that maintain the general template of a genuine question but are \textit{\textbf{factually incorrect}}. We therefore assume a ratio of 9:1 correct:incorrect tasks and we use \texttt{GPT-4} to generate variations per template for a total of 1,000 tasks. To allow \texttt{GPT-4} to ``programmatically'' select from real data combinations, we provide in-context prompt with dataset descriptions, \textit{e.g.}, SQL schemas with all eligible category names in the \texttt{xview1} database.

\begin{tcolorbox}[title=Reference Question with Paraphrased and Perturbed Variations, colback=gray!20, colframe=gray!75, rounded corners, sharp corners=northeast, sharp corners=southwest]
\footnotesize
\texttt{\textbf{Reference Q:} Use the YOLO detector to detect fishing vessels in xview3 images around Ancona. Plot them on the map.\\
\textbf{Paraphrased Q:} Use RetinaNet to find yacht detections in xview1 images around Barbados, and show them on the map.\\
\textbf{Perturbed Q:} Use NoNet to find Zeppelins in images around the mythical city of Atlantis.}
\end{tcolorbox}

\textit{3. Human-guided Ground Truth Generation}: last, to generate the ground truth answers and tool-set solutions, we task \texttt{GPT-4} to solve each question using the available platform tools. To guide the process, we leverage the Reference Templates (questions and solutions) and we augment the LLM by dynamically retrieving similarly correct examples via RAG~\citep{gao2024retrievalaugmented}. This allows us to accelerate the process, while ensure the human-on-the-loop to validate the overall correctness.

\textbf{Metrics}: Unlike existing VQA-based benchmarks, we consider a a comprehensive set of metrics that capture the LLM's ability for effective tool-calling and reasoning:

\vspace{-4pt}
\textit{\textbf{a.} Success rate}: the ratio of successfully completed tasks across the entire benchmark. Each task is consider to be completed correctly when the final platform state $S$ matches the $\tilde{S}$ ground-truth. This ratio informs us of the degree to which the agent is able to complete tasks, irrespective of whether it took incorrect or unnecessary intermediate steps. 

\vspace{-4pt}
\textit{\textbf{b.} Correctness ratio}: the ratio of correct function-call operations across the benchmark. Given a ground-truth tool-set $\tilde{T}$ and an LLM solution $T$, we track \textit{all} applicable LLM error-types as defined in~\citep{zhuang2023toolqa} (\textit{i.e.}, ``Infeasible Action'',  ``Function Error'', ``Argument Error'', ``Incorrect Data Source'', and ``Omitted Function''). Given the total number of errors and ground-truth tools, we compute the correctness ratio $R_{correct} = \max(0, 1- N_{errors}/N_{tools})$~\citep{maini2024tofu}. This metric captures how likely it is for the agent to invoke the correct functions in the expected order. 

\vspace{-4pt}
\textit{\textbf{c.} ROUGE score}: we use the ROUGE-L recall score~\citep{lin04rouge} to compare model answers $a$ with the ground truth $\tilde{a}$ to assess the ability of the agent to reply to the task at hand. 

\vspace{-4pt}
\textit{\textbf{d.} Cost (Tokens)}: we compute the average number of tokens per task over the entire benchmark.

\vspace{-4pt}
\textit{\textbf{e.} (Detection) Recall}: over the entire benchmark, we assess the agents ability to correctly return detection tasks by calculating the overall recall $R$ (\textit{i.e.}, detections returned by the method against ``gold'' ground-truths from oracle detectors).

\section{Experiments}

In the scope of this analysis, we run various prompting techniques from literature: Chain-of-Thought~\citep{wei2023chainofthought}, (MM-)ReAct~\citep{yao2023react, yang2023mmreact}, and Chameleon~\citep{lu2023chameleon}. We leave more advanced prompting strategies for future investigation. Our baselines language models include GPT-4 Turbo (\texttt{gpt-4-0125-preview}) and GPT-3.5 Turbo (\texttt{gpt-3.5-turbo-1106}).

\begin{table}[t!]
\caption{Performance of different agents on \texttt{GeoLLM-QA-1k}.}
\label{tab:results_easy}
\begin{center}
\resizebox{1.0\textwidth}{!}{%
\begin{tabular}{lccccc}
~ & \textbf{Success} &  \textbf{Correctness} &  \textbf{ROUGE}& \textbf{Det.} & \textbf{Avg. Tokens}$\downarrow$ \\
~ & \textbf{Rate}$\uparrow$ &  \textbf{Rate}$\uparrow$ &  \textbf{-L}$\uparrow$ & \textbf{Recall}$\uparrow$ & \textbf{/Task}$\downarrow$ \\\hline \\
\textbf{\textit{GPT-3.5 Turbo (0125)}} & & & & & \\ 
CoT~\citep{wei2023chainofthought} Zero-Shot & 30.74\% & 80.67\% & 21.42\% & 91.92\% & 7.4k \\
CoT~\citep{wei2023chainofthought} Few-Shot & 31.65\%  &  89.55\% & 22.05\% & 71.17\% & 9.3k \\
Chameleon~\citep{lu2023chameleon} Zero-Shot & 23.69\%  &  79.88\% & 23.29\% & 89.73\% & 12.1k \\
Chameleon~\citep{lu2023chameleon} Few-Shot & 26.74\%  &  85.70\% & 24.30\% & 96.18\% & 12.9k \\
ReAct~\citep{yao2023react} Zero-Shot & 30.70\%  &  86.26\% & 22.31\% & 77.17\% & 7.5k \\
ReAct~\citep{yao2023react}  Few-Shot & 32.95\%  &  89.35\% & 26.06\% & 91.78\% & 11.1k \\\\
\textbf{\textit{GPT-4 Turbo (0125)}} & & & & & \\ 
CoT~\citep{wei2023chainofthought} Zero-Shot & 34.99\%  &  94.59\% & 26.82\% & 85.81\% & 8.7k \\
CoT~\citep{wei2023chainofthought} Few-Shot & 33.35\%  &  94.93\% & 27.09\% & 93.33\% & 9.2k \\
Chameleon~\citep{lu2023chameleon} Zero-Shot & 29.44\%  &  83.49\% & 21.57\% & 88.88\% & 12.5k \\
Chameleon~\citep{lu2023chameleon} Few-Shot & 31.18\%  &  89.59\% & 22.56\% & 90.41\% & 13.1k \\
ReAct~\citep{yao2023react} Zero-Shot & 33.52\%  &  94.85\% & 27.82\% & 87.77\% & 9.5k \\
ReAct~\citep{yao2023react}  Few-Shot & 33.39\%  &  94.98\% & 27.75\% & 96.73\% & 11.6k \\
\end{tabular}
}
\end{center}
\end{table}

Tab.~\ref{tab:results_easy} summarizes our findings. The recent GPT-4 Turbo release exhibits impressive function-calling capabilities, while in terms of methods, CoT and ReAct outperform Chameleon in both correctness and success rates, while being more token efficient. With respect to other metrics, ROUGE-L shows the limitations of text-based scores, as it has been reported by recent work on foundation models comparing closed- and open-vocabulary answers~\citep{OpenEQA2023}. That is, the distribution of LLM answers is heavily dependent on the prompting method. For instance, answers generated by GPT-3.5 might artificially penalize a different response style by Chameleon if treated ground-truths (\textit{e.g.}, ``There are five airplanes'' \textit{vs}. ``This image contains 5 planes'' can result in lower scores despite conveying the same fact). Last, we observe that detection-related metrics, as captured by \textit{recall}, do not necessarily correlate with agent performance. All these findings confirm that, unlike existing RS benchmarks that mainly report detection results or captioning-related scores, a more comprehensive evaluation is required to assess agent performance. 

\begin{figure}[h!]
    \centering    
    \includegraphics[width=1.0\linewidth]{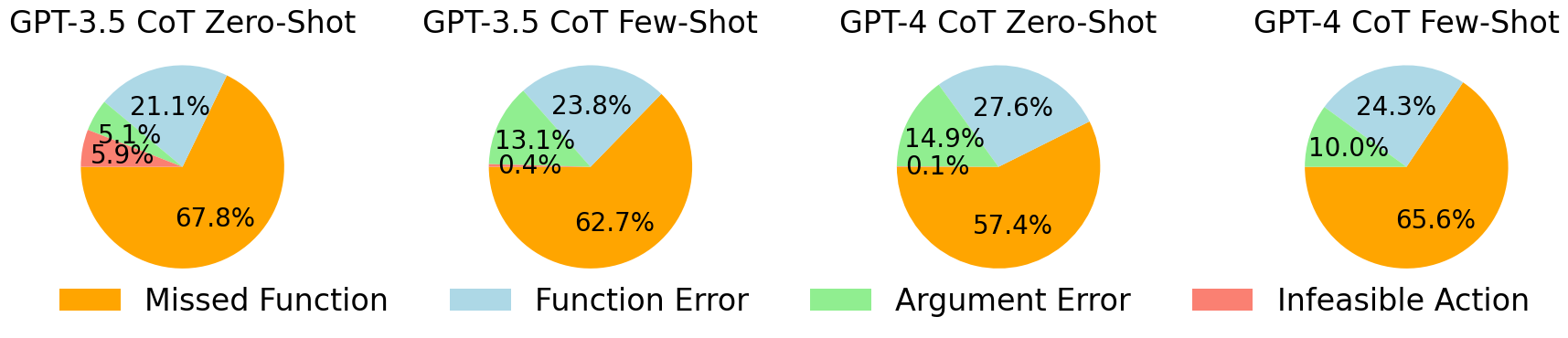}
    \caption{GPT-3.5 \textit{vs}. GPT-4 error analysis for CoT prompting.}
    \label{fig:errorann}
\end{figure}

Fig.~\ref{fig:errorann} shows the error types for CoT on GPT-3.5 and GPT-4, in both zero-shot and few-shot scenarios. The most common, ``Missed Function'' (where the agent omits necessary tool calls regardless of the approach used) accounts for more than half of all errors. We expect that dynamic/RAG-augmented~\citep{srinivasan2023nexusraven} prompting should improve agent performance by addressing such failures. Last, the consistent distribution across different cases implies that these issues are not method-specific but rather inherent to the current GPT capabilities.

\section{Conclusion and Future Work}

We presented \texttt{GeoLLM-QA}, a benchmark of realistic \textit{user-grounded} tasks aimed at assessing the capabilities of tool-augmented LLMs in geospatial applications. Our hope is that this benchmarking suite will spur the development of new agents that advance the state of the art in remote sensing platforms. To this end, we would like to highlight some particularly exciting and promising areas for future work that we have identified through our research and that we are actively investigating. First, recent advances in multimodal modeling show improved performance compared to MM-ReAct-like prompting. We are currently extending our benchmark to flexibly incorporate open-source GPT-V model families, such as mini-GPT~\citep{zhu2023minigpt4, chen2023minigptv2}. Additionally, we are expanding our analysis to replace oracle detectors with state-of-the-art models~\citep{jian2023stable}, to explore how agent errors interact with suboptimal detector performance.

Moreover, a primary bottleneck that we have encountered with our approach, which is common in related work~\citep{zhan2024skyeyegpt}, is the overhead of human-guided template generation. In our most recent study~\citep{singh2024geoengine}, we demonstrate that by adopting engine-based benchmarking methodologies~\citep{zhou2023webarena} in the remote sensing domain, we can leverage fully GPT-driven template and ground-truth generation to minimize human-in-the-loop overhead. Lastly, by considering cost- and system-related aspects, our analysis has yielded interesting insights regarding optimizing the overall agent-platform implementation. Our ongoing explorations include methods to improve performance by leveraging state-of-the-art LLM caching and compression techniques~\citep{jiang2023llmlingua, fore2024geckopt}.

\bibliography{main}
\bibliographystyle{iclr2024_conference}

\end{document}